\documentclass[journal]{IEEEtran}
\ifCLASSINFOpdf
  \usepackage[pdftex]{graphicx}
\else
\fi

\usepackage{diagbox}
\usepackage{adjustbox} 
\begin{document}
%
\title{Transition Subspace Learning based Least Squares Regression for Image Classification}
%
%
%

\author{Zhe~Chen,~\IEEEmembership{}
        Xiao-Jun~Wu$^*$,~\IEEEmembership{}
        and~Josef~Kittler,~\IEEEmembership{Life~Member,~IEEE}
\IEEEcompsocitemizethanks{\IEEEcompsocthanksitem $*$ Corresponding author. \protect
\IEEEcompsocthanksitem Zhe Chen and Xiao-Jun Wu are with the School of Internet of Things, Jiangnan University, Wuxi 214122, China.\protect\\
E-mail: 7181905012@stu.jiangnan.edu.cn, wu\_xiaojun@jiangnan.edu.cn
\IEEEcompsocthanksitem Josef Kittler is with the Centre for Vision, Speech and Signal Processing, University of Surrey, Guildford GU2 7XH, U.K. \protect\\
E-mail: j.kittler@surrey.ac.uk }
\thanks{}}

%
%

\markboth{}%
{Shell \MakeLowercase{\textit{et al.}}: Bare Demo of IEEEtran.cls for IEEE Journals}
%



\maketitle

\begin{abstract}
Only learning one projection matrix from original samples to the corresponding binary labels is too strict and will consequentlly lose some intrinsic geometric structures of data. In this paper, we propose a novel transition subspace learning based least squares regression (TSL-LSR) model for multicategory image classification. The main idea of TSL-LSR is to learn a transition subspace between the original samples and binary labels to alleviate the problem of overfitting caused by strict projection learning. Moreover, in order to reflect the underlying low-rank structure of transition matrix and learn more discriminative projection matrix, a low-rank constraint is added to the transition subspace. Experimental results on several image datasets demonstrate the effectiveness of the proposed TSL-LSR model in comparison with state-of-the-art algorithms. This paper is under consideration at Pattern Recognition Letters.
\end{abstract}


%
\IEEEpeerreviewmaketitle

\section{Introduction}
%
%
%
%
\IEEEPARstart{L}{east} squares regression (LSR) is a very popular tool in the field of pattern recognition, becasuse of its computational efficiency and mathematical tractability. Many modified models, including LASSO regression \cite{LASSO}, partial LSR \cite{PLSR}, least-square support vector machine \cite{SVM}, kernel ridge regression \cite{KRR}, weight LSR \cite{WLSR}, were proposed for classification tasks. Besides, some representation based classification algorithms, such as sparse representation based classification (SRC) \cite{SRC}, linear regression based classification (LRC) \cite{LRC}, collaborative representation based classification (CRC) \cite{CRC} and probabilistic CRC (ProCRC) \cite{ProCRC}, are also calculated under the LSR model. These algorithms have achieved varying degrees of success in improving classification accuracy.

Consider $n$ training samples $\{x_1,x_2,...,x_n\}$ from $c$ classes, where $x_i\in R^d$ denotes a sample vector. $d$ is the dimensionality of the sample. If collecting these samples as a training matrix $X=[x_1,x_2,...,x_n]\in R^{d\times n}$, the standard LSR model can be defined as follows
$$\min_W\|WX-H\|_F^2+\lambda\|W\|_F^2  \eqno{(1)}$$
where $\lambda$ is a regularization parameter and $W\in R^{c\times d}$ is the projection matrix which to be learned. $Y=[y_1,y_2,...,y_n]\in R^{c\times n}\ (c\geq 2)$ is the binary label matrix. The $i$th column of $H$, i.e., $h_i=[0,0,...,0,1,0,...,0]^T \in R^c$, is the label vector of sample $x_i$. Suppose $x_i$ is from the $j$th class $(j=1,2,...,c)$, then only the $j$th element of $h_i$ is equal to 1 and all the others are 0. Obviously, problem (1) has a closed-form solution $\hat W=HX^T(XX^T+\lambda I)^{-1}$. For a given test sample $y\in R^d$, LSR predicts its label as $l=argmax_i(Wy)_i$, where $(Wy)_i$ is the $i$th entry of $Wy$.

In recent years, researchers developing LSR have focused more on learning relaxed regression targets to replace zero-one labels. For example, Xiang et al. \cite{DLSR} presented a discriminative least squares regression (DLSR) model by utilizing a technique called $\varepsilon$-dragging. The idea of DLSR was to enlarge the margins between the true and the false classes as much as possible, after the original samples are projected into corresponding label space, which intuitively facilitates classification. 
Retargeted LSR (ReLSR) \cite{ReLSR} directly learned the regression targets from data which can guarantee all samples are correctly classified with the large margins. Wang et al. \cite{GReLSR} proposed a new groupwise ReLSR (GReLSR) model by introducing a groupwise regularization term to encourage the within-class samples have similar translation values. 

However, directly minimizing the regression error between the projection features and labels is too restrictive. Only one projection matrix is not enough to contain sufficient discriminative information. Besides, both $\varepsilon$-dragging and margin constraint techniques can also enlarge the distances between the within-class regression targets. In addition to learning relaxed targets, RLSL \cite{RLSL} proposed to learn a latent feature subspace that can be regarded as a intermediate between the original samples and binary labels. Nevertheless, RLSL did not take into account the structural characteristics of learned latent subspace. 

In this paper, a novel transition subspace learning based LSR (TSL-LSR) model is proposed for multiclass classification. The main advantage of TSL-LSR is the learning of transition subspace which can preserve more underlying structural information in the learned projection. Specifically, the contributions of TSL-LSR can be highlighted as follows

(1) We propose to learn a transition subspace to avoid the problem of over-fitting, which is more flexible than learning projection from samples to zero-one labels directly. 

(2) TSL-LSR first transforms the original samples into a transition subspace, then transforms the transition subspace into the space of binary labels. Hence, there are two projection matrices to be learned in the TSL-LSR model and both of these two matrices are used for classification. 

(3) To guarantee consistency and global optimum of transformation learning, two projection matrices are learned in a joint framework.

(4) A low-rank constraint is imposed on the transition matrix to capture the underlying feature structures (low-rank structure) of different classes.

(5) The low-rank transition subspace can also be extended to the slack targets based LSR models which is helpful to learn similar and compact within-class regression targets. 

\section{Transition Subspace Learning based Least Squares Regression (TSL-LSR)}
\subsection{The Model of TSL-LSR}
Since binary labels already have enough discriminability for classification, TSL-LSR still uses the zero-one labels as the final regression targets. But unlike DLSR, ReLSR and GReLSR, TSL-LSR learns discriminative projections by introducing a low-rank transition subspace to avoid the loss of structural information, rather than relaxing the binary regression targets. The model of TSL-LSR can be formulated as
\setcounter{equation}{1} 
\begin{eqnarray}
\min_{W,Q,\Omega}\frac{1}{2}\|WX-\Omega\|_F^2+\alpha\|\Omega\|_*+\frac{\beta}{2}\|Q\Omega-H\|_F^2+ \nonumber \\ 
\frac{\lambda_1}{2}\|W\|_F^2+\frac{\lambda_2}{2}\|Q\|_F^2
\end{eqnarray}
where $\alpha$, $\beta$, $\lambda_1$ and $\lambda_2$ are positive regularization parameters. $W\in R^{p\times d}$, $Q\in R^{c\times p}$ and $\Omega\in R^{p\times n}$ are variables which need to be optimized. $\Omega$ is the transition matrix and $p$ is the dimensionality of transition subspace. $W$ and $Q$ are two projection matrices.  $\|\bullet\|_*$ is the nuclear norm operator (the sum of matrix singular values) and $\|\Omega\|_*$ denotes the low-rank constraint on matrix $\Omega$.  

The consequence of introducing  the transitional transformation space, $\Omega$, is that TSL-LSR  must learn two projection matrices in one model. However, this is more flexible than learning one projection matrix. The first projection matrix, $W$, is used to transform the original samples into the transition subspace, and the second, $Q$, is used to transform the transition subspace into the space of binary labels. The reasons for adding a low-rank constraint on transition subspace $\Omega$ can be summarized as follows

(1) The final regression targets, i.e. label matrix $H$, are low-rank (rank=$c$), thus it is reasonable to assume the transition space is also low-rank.

(2) For real-world image classification tasks, images are often collected in realistic conditions, so that they are subject to noise, which has an adverse effect on classification. Thus we assume that the features obtained after the first-step projection, i.e. $WX$, are heterogeneous. We try to recover a low-rank subspace from the corrupted features based on the assumption that the clean data structures are approximately drawn from a low-rank subspace. As a result, more useful structure information of images can be captured during the transformation learning process. The proposed learning framework (2) is illustrated in Fig. 1. As shown in Fig. 1, we find that the features extracted by our TSL-LSR model include two parts: the first-step features $\Omega$ and the second-step features $Q\Omega$.

\begin{figure}[!h]
\centering
\includegraphics[scale=0.46]{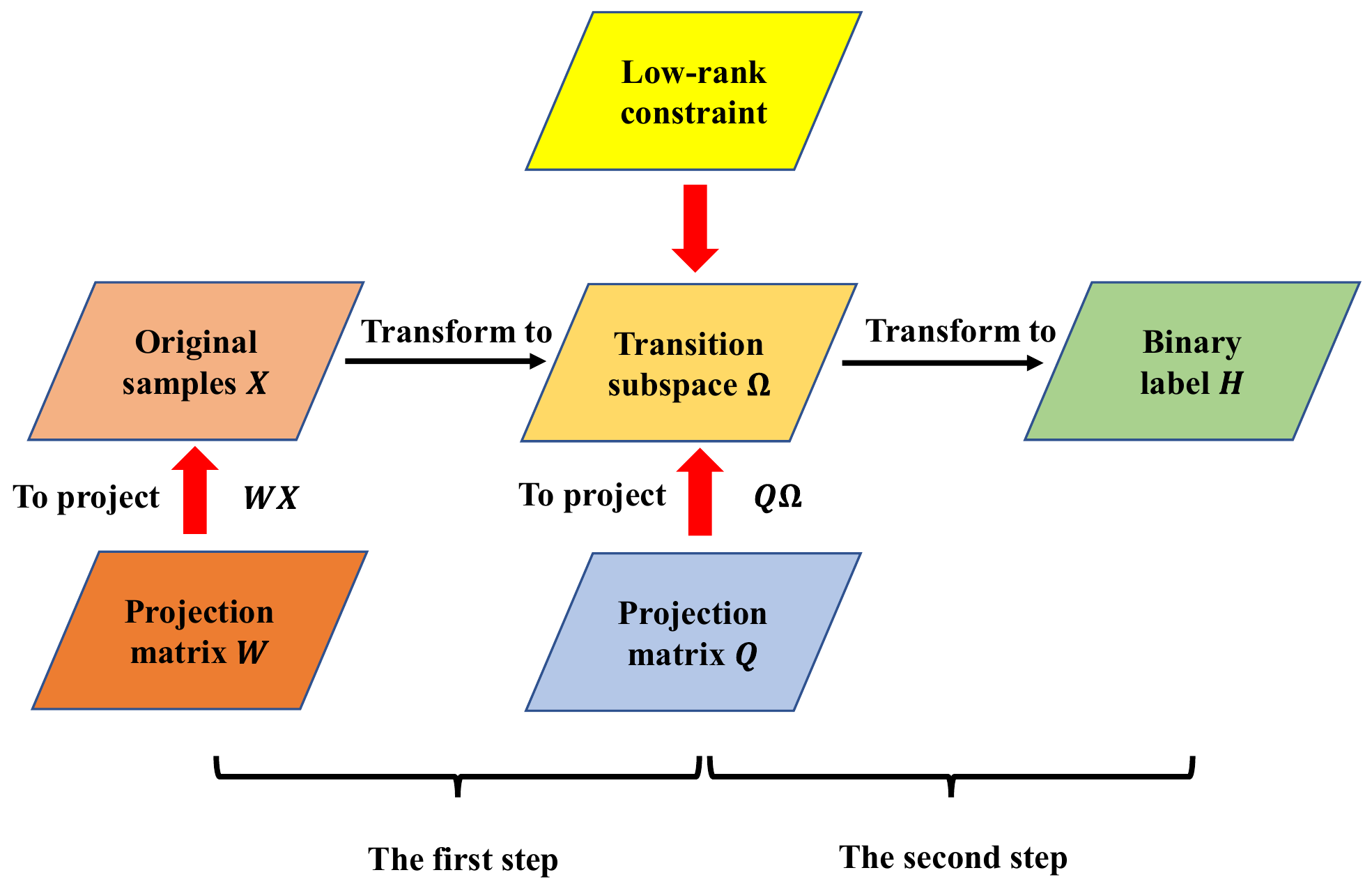}
\caption{Illustration of the proposed TSL-LSR model. $WX$ and $Q\Omega$ denote the features of the first-step projection and the second-step projection, respectively.}
\end{figure}

\subsection{Optimization of TSL-LSR}
The objective function in (2)
cannot be directly optimized because
the variables (i.e, $W$, $Q$ and $\Omega$ ) are interdependent. Therefore, we use the alternating direction multipliers method (ADMM) \cite{ADMM} to solve the optimization problem. We first introduce an auxiliary variable $P$ to make problem (2) separable and give its augmented Lagrangian function as
\setcounter{equation}{2}  
\begin{eqnarray}
L(W,Q,\Omega,P,Y)=\frac{1}{2}\|WX-\Omega\|_F^2+\alpha\|P\|_*+ \nonumber \\ 
\frac{\beta}{2}\|Q\Omega-H\|_F^2+ 
\frac{\lambda_1}{2}\|W\|_F^2+\frac{\lambda_2}{2}\|Q\|_F^2+\nonumber \\ 
\frac{\mu}{2}\|\Omega-P+\frac{Y}{\mu}\|_F^2
\end{eqnarray}
where $Y$ is the Lagrangian multiplier, $\mu>0$ is the penalty parameter. Each variable, such as $W$, $Q$, $\Omega$ and $P$, is updated with other variables fixed.

\noindent\textbf{Update $W$:} By fixing variables $Q$, $\Omega$ and $P$, $W$ can be obtained by minimizing the following problem
$$L(W)=\frac{1}{2}\|WX-\Omega\|_F^2+\frac{\lambda_1}{2}\|W\|_F^2 \eqno{(4)}$$

We set the derivative of $L(W)$ with respect to $W$ to zero, and obtain the following closed-form solution

$$W=\Omega X^T(XX^T+\lambda_1I)^{-1} \eqno{(5)}$$

\noindent\textbf{Update $Q$:} $Q$ can be obtained by minimizing the following problem
$$L(Q)=\frac{\beta}{2}\|Q\Omega-H\|_F^2+\frac{\lambda_2}{2}\|Q\|_F^2 \eqno{(6)}$$
which has a closed-form solution as
$$Q=\beta H\Omega^T(\beta \Omega\Omega^T+\lambda_2I)^{-1}\eqno{(7)}$$

\noindent\textbf{Update $\Omega$:} $\Omega$ can be obtained by minimizing the following problem
$$L(\Omega)=\frac{1}{2}\|WX-\Omega\|_F^2+\frac{\beta}{2}\|Q\Omega-H\|_F^2+\frac{\mu}{2}\|\Omega-P+\frac{Y}{\mu}\|_F^2 \eqno{(8)}$$
Likewise, $\Omega$ has a closed-form solution
$$\Omega=[(\mu+1)I+\beta Q^TQ]^{-1}(WX+\beta Q^TH+\mu P-Y) \eqno{(9)}$$

\noindent\textbf{Update $P$:} $P$ can be obtained by minimizing the following problem
$$L(P)=\alpha\|P\|_*+\frac{\mu}{2}\|\Omega-P+\frac{Y}{\mu}\|_F^2 \eqno{(10)}$$

Formula (10) can be optimized by the singular value thresholding algorithm \cite{SVD}. The optimal solution of (10) is 
$$P=I_{\frac{\alpha}{\mu}}(\Omega+\frac{Y}{\mu}) \eqno{(11)}$$
where $I_\zeta(\Theta)$ is the singular value shrinkage operator. The complete optimization procedures are summarized in Algorithm 1.
\begin{table}[!ht]
\rule[0.1cm]{8.8cm}{1.5pt}
\leftline {\textbf {Algorithm 1.} Solving TSL-LSR by ADMM}\\
\rule[0.1cm]{8.8cm}{1.5pt}
\textbf{Input:} Normalized training samples $X$ and its label matrix $H$; Parameters $\alpha, \beta, \lambda_1, \lambda_2$ .

\textbf{Initialization:} $W=Q=P=\textbf0$, $\Omega=H$, $Y=\textbf0$, $\mu_{max}=10^{8}$, $tol=10^{-6}$, $\mu=10^{-5}$, $\rho=1.1$.\\
\textbf{While} not converged do:
 \begin{enumerate}
\item Update $W$, $Q$, $\Omega$ and $P$ one by one.
\item Update Lagrange multipliers $Y$ as
$$Y=Y+\mu(\Omega-P). \eqno{(12)}$$
\item Update penalty parameter $\mu$ as
$$\mu=min(\mu_{max}, \rho \mu). \eqno{(13)}$$
\item Check convergence: 
$$if \ \|\Omega-P\|_{\infty}\leq tol.$$
\end{enumerate}
\textbf{End While}\\
\textbf{Output:} $W$ and $Q$ \\
\rule[0.1cm]{8.8cm}{1.5pt}
\end{table}

Next, we analyze the computational complexity of Algorithm 1. Following \cite{IMSC}, the main time-consuming steps of Algorithm 1 are

(1) Matrix inverse in Eq. (5), (7), and (9).

(2) Singular value decomposition in Eq. (11).

The complexity of pre-computing $X^T(XX^T+\lambda_1I)^{-1}$ in Eq. (5) is $O(d^3)$.  The complexity of computing each of $(\beta \Omega\Omega^T+\lambda_2I)^{-1}$ in Eq. (7) and $[(\mu+1)I+\beta Q^TQ]^{-1}$ in Eq. (9) is $O(c^3)$. The complexity of singular value decomposition in Eq. (11) is $O(n^3)$. Thus the final time complexity for Algorithm 1 is about $O(d^3 + \tau (c^3+n^3))$, where $\tau$ is the number of iterations.

\subsection{Classification}
Once the optimal projection matrices $W$ and $Q$ are obtained, we
can use them to classify test samples. Given a new test sample $y\in R^d$, its regression  is $QWy$. Then, the nearest-neighbor (NN) classifier is used to predict the label of $y$.

\section{Experiments}
We compare the proposed TSL-LSR model with four latest LSR model based classification methods, including DLSR \cite{DLSR}, ReLSR \cite{ReLSR}, GReLSR \cite{GReLSR}, RLSL \cite{RLSL}, and three representation based classification methods, including LRC \cite{LRC}, CRC \cite{CRC}, and ProCRC \cite{ProCRC}, on a range of different datasets. For TSL-LSR, DLSR, ReLSR, GReLSR and RLSL, we use the NN classifier. The used datasets consists of two types:
(1) Face: the AR \cite{AR}, CMU PIE \cite{CMU_PIE} and Feret \cite{Feret} datasets; (2) Object: the COIL-20 \cite{COIL} dataset. For each dataset, we randomly select several images of each class for training, and the remaining images are used for testing. We repeat all the experiments ten times and report the mean classification results (mean$\pm$std). The brief description of these datasets are shown in Table I. 

\begin{table}[htbp]
\setlength{\tabcolsep}{3.8pt}
\renewcommand{\arraystretch}{1.5}
\caption{Brief description of the used five datasets.}
\centering
\begin{tabular}{|c|c|c|c|c|}
\hline
\diagbox{Dataset}{Info.} & Classes & Features & Total Num. & Training Num. \\
\hline
AR & 100 & 540 & 2600 & 1000 \\
\hline
CMU PIE & 68 & 1024 & 11554 & 680\\
\hline
Feret & 200 & 1600 & 1400 & 800\\
\hline
COIL-20 & 20 & 1024 & 1440 & 200\\
\hline
\end{tabular}
\end{table}

\subsection{Classification results on different datasets}
We first need to determine the value of $p$, where $p$ is the row dimensionality of transition matrix $\Omega$. In fact, it is very difficult to tune its value, because $p$ could be $(0, +\infty)$. From \cite{p}, we know $p$ can be set to around $c$, where $c$ is the number of classes. Fig. 2 presents the classification accuracies (\%) versus the value of $p$ on two face datasets. We can see that the change in accuracy is not obvious while $p>c$ and the peak is achieved if $p$ is approximately equal to $c$. Therefore, in our experiments, we directly fix $p=c$ on all datasets.

\begin{figure}[]
\centering
\includegraphics[scale=0.29]{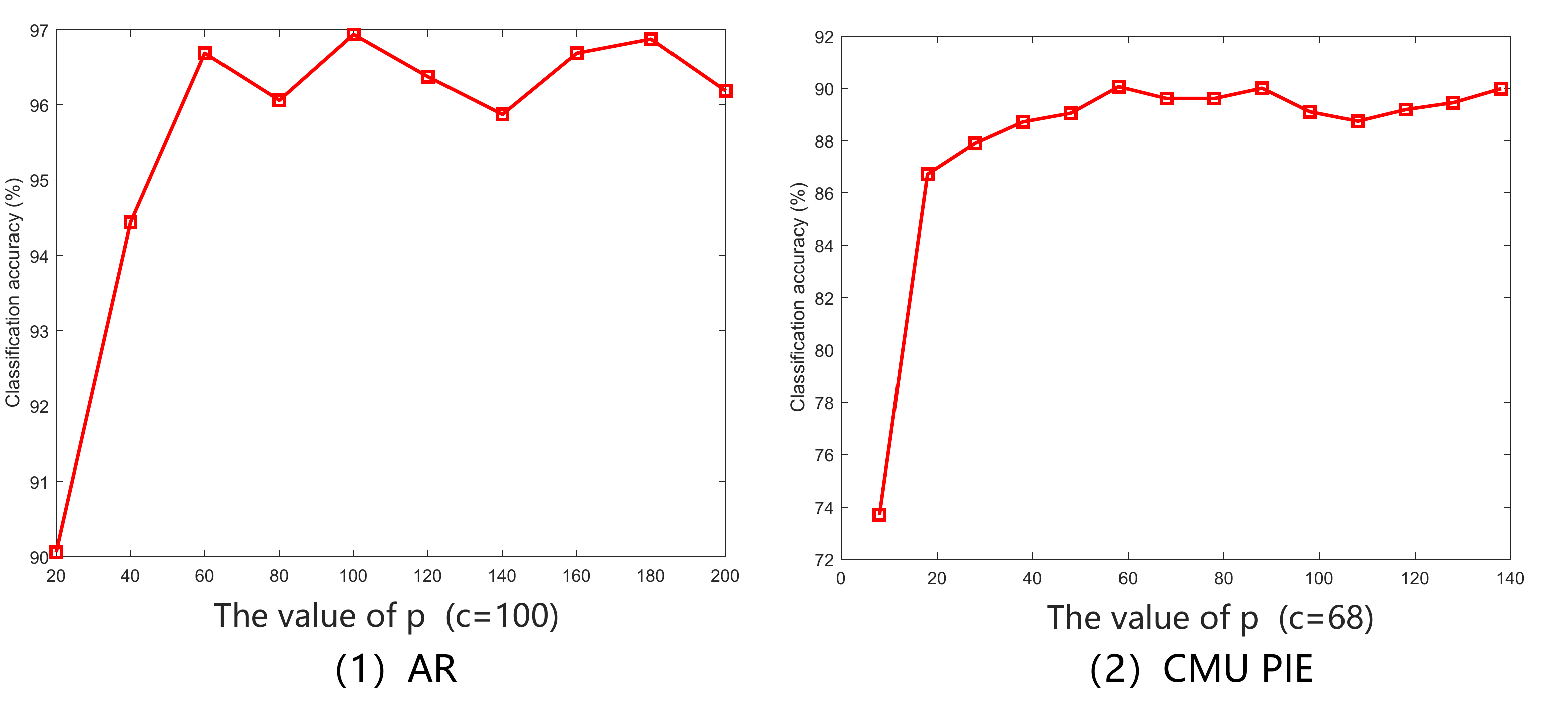}
\caption{Classification accuracies (\%) versus the dimension of transition space $\Omega$ on (1) AR and (2) CMU PIE datasets, in which we randomly select 10 images per class for training and the remaining images are used as testing set.}
\end{figure}

The comparative classification results on five datasets are shown in Table II. As shown in Table II, our TSL-LSR model consistently achieves better accuracies than the other algorithms, including the latest two algorithms, such as GReLSR and RLSL. This is mainly because both DLSR, ReLSR and GReLSR algorithms focus on learning slack regression targets without guarding against the problem of over-fitting. In contrast, TSL-LSR introduces a low-rank transition subspace to alleviate the structural information loss caused by restrictive matrix projection. Its learned two projection matrices have a greater capacity to capture the discriminative information conveyed by the data during projection learning. To further validate that whether the learned two projections from TSL-LSR model can capture discriminative features from original samples, we use the t-SNE algorithm \cite{t-SNE} to visualize the distribution of the extracted features. From Fig. 3, we can find that TSL-LSR correctly distributes all the samples into their own subspace and the distribution of intra-class samples are very compact which indicates that the extracted features perform ideal inter-class separability and intra-class compactness. This also demonstrates that the transition subspace learning is beneficial for classification.

\begin{table}[!h]
\setlength{\tabcolsep}{3pt}
\renewcommand{\arraystretch}{1.5}
\caption{Classification accuracies (\%) of different algorithms on different datasets.}
\centering
\begin{tabular}{|c|c|c|c|c|c|}
\hline
Algorithms & AR & CMU PIE & Feret & COIL-20 \\
\hline
LRC\cite{LRC} & 74.12$\pm$1.50 & 75.67$\pm$1.01 & 46.58$\pm$1.33 & 92.30$\pm$1.15   \\
\hline
CRC\cite{CRC} & 93.36$\pm$0.53 & 86.39$\pm$0.60 & 57.07$\pm$1.79 & 89.09$\pm$1.48 \\
\hline
ProCRC\cite{ProCRC} & 95.28$\pm$0.41 & 89.00$\pm$0.37 & 64.40$\pm$2.54 & 90.61$\pm$0.95 \\
\hline
DLSR\cite{DLSR} & 93.79$\pm$0.50 & 87.54$\pm$0.79 & 71.15$\pm$1.27 & 93.27$\pm$1.43  \\
\hline
ReLSR\cite{ReLSR} & 94.53$\pm$0.56 & 88.18$\pm$0.79 & 72.98$\pm$2.19 & 93.65$\pm$1.94   \\
\hline
GReLSR\cite{GReLSR} & 95.18$\pm$0.74 & 86.88$\pm$0.72 & 70.38$\pm$2.14 & 90.98$\pm$1.62  \\
\hline
RLSL\cite{RLSL} & 94.21$\pm$0.35 & 87.70$\pm$0.63 & 68.33$\pm$1.57 & 93.75$\pm$1.87  \\
\hline
TSL-LSR (ours) & \textbf{96.34$\pm$0.43} & \textbf{89.92$\pm$0.35} & \textbf{85.73$\pm$1.39} & \textbf{94.34$\pm$1.02} \\
\hline
\end{tabular}
\end{table}

\begin{figure}[!h]
\centering
\includegraphics[scale=0.35]{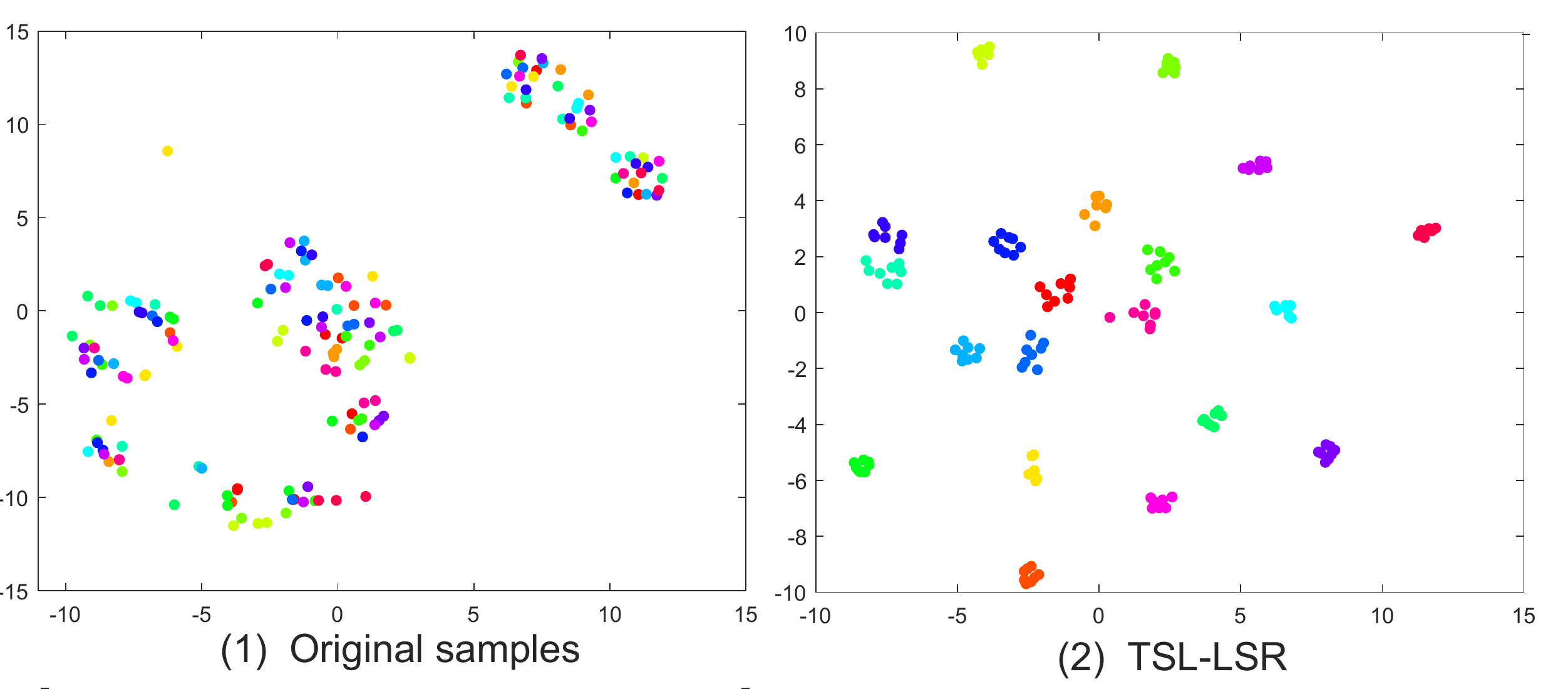}
\caption{t-SNE visualization results of the features extracted by TSL-LSR on the AR dataset, where random 8 images of each person and the first 20 persons are selected for validation. Subfigure (1) and (2) correspond to original features and TSL-LSR features ($Q\Omega$), respectively. }
\end{figure}

\subsection{Convergence Validation}
Based on the optimization procedures in Section II(B), it is easy to prove that the proposed TSL-LSR model is convex with respect to each variable. In this section, we validate the convergence of Algorithm 1 on two datasets. The convergence results are shown in Fig. 4. We can see that Algorithm 1 converges very well, with the value of objective function of TSL-LSR monotonically decreasing with the increasing number of iterations. This confirms the effectiveness of the adopted optimization algorithm.

\begin{figure}[!h]
\centering
\includegraphics[scale=0.55]{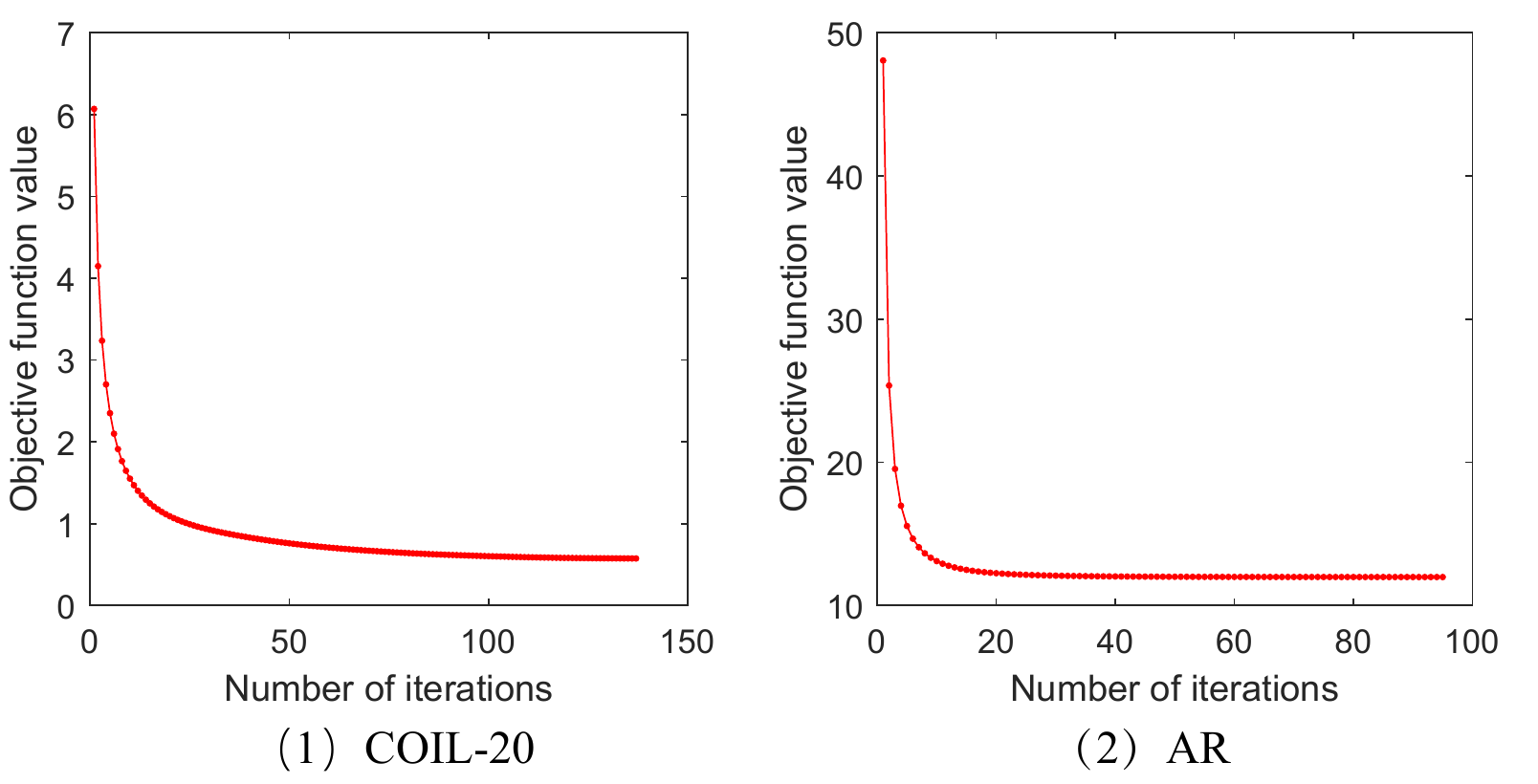}
\caption{Convergence curves of TSL-LSR on (1) COIL-20 and (2) AR datasets.}
\end{figure}

\subsection{Parameter Sensitivity}
In this section, we test the parameter sensitivity of TSL-LSR. TSL-LSR has four parameters to be tuned in our experiments. The parameters $\lambda_1$ and $\lambda_2$ are both set to 0.01, so we just focus on selecting the values of parameters $\alpha$ and $\beta$ from the candidate set
$\{0.001, 0.005, 0.01, 0.05, 0.1, 0.5, 1\}$. The classification accuracy as a function of different parameter values on the four datasets are shown in Fig. 5.  It is apparent that the classification accuracy of TSL-LSR is not very sensitive to the values of $\alpha$ and $\beta$.

\begin{figure}[!h]
\centering
\includegraphics[scale=0.45]{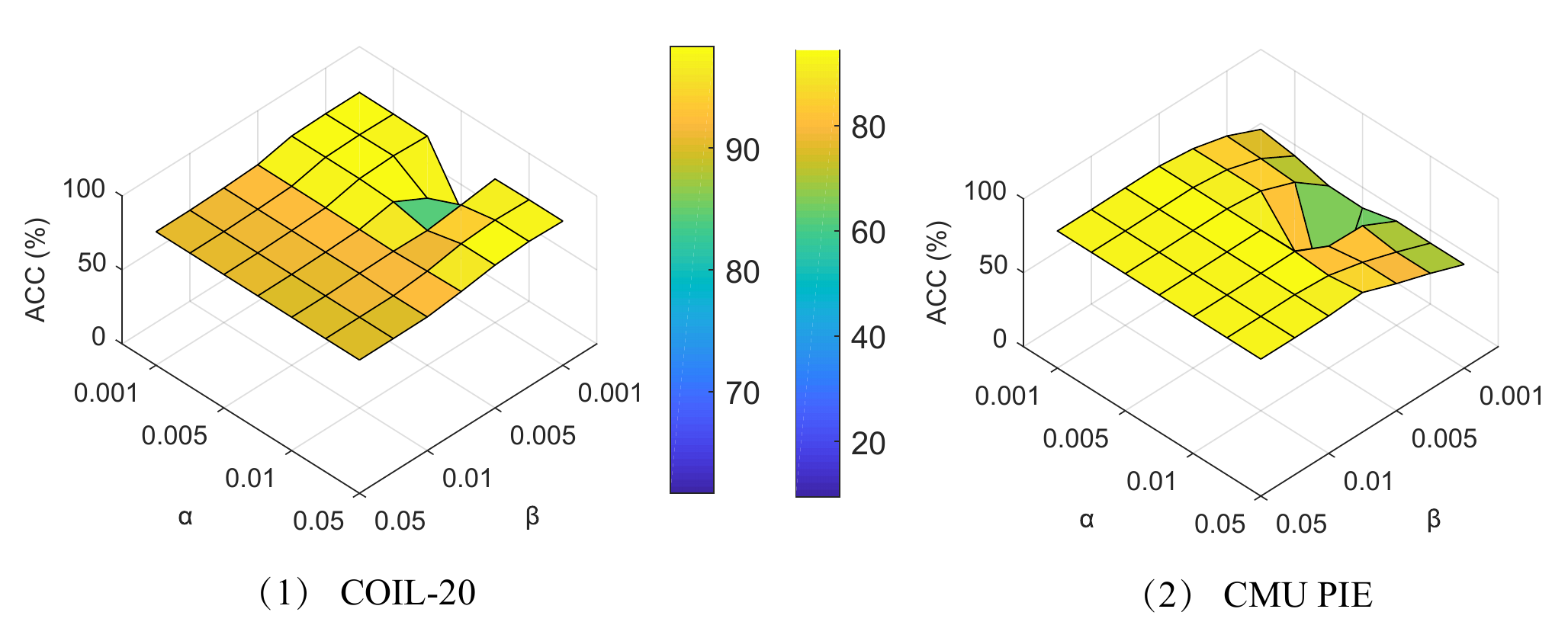}
\caption{The performance evaluation (\%) of TSL-LSR versus parameters $\alpha$ and $\beta$ on (1) COIL-20 and (2) CMU PIE datasets.}
\end{figure}

\section{Conclusion}
In this paper, an effective transition subspace learning based least squares regression model (TSL-LSR) 
is proposed for multicategory image classification. Different from traditional LSR based regression models, which directly learn projection from original samples to corresponding label subspace, TSL-LSR tries to learn a low-rank transition subspace to avoid the problem of overfitting caused by restrictive projection learning. Moreover, TSL-LSR imposes a low-rank constraint on the transition matrix to learn more underlying structures of data. Two discriminative projection matrices are learned for classification. Extensive experiments demonstrate the effectiveness of the proposed method.


%





\ifCLASSOPTIONcaptionsoff
  \newpage
\fi


\begin{thebibliography}{1}

\bibitem{LASSO}
R. Tibshirani, "Regression shrinkage and selection via the lasso," J. Roy.
Statist. Soc. B (Methodol.), vol. 58, no. 1, pp. 267-288, 1996.

\bibitem{PLSR}
S. Wold, H. Ruhe, H. Wold, and W. Dunn, "The collinearity problem in
linear regression. the partial least squares (PLS) approach to generalized
inverses," J. Sci. Stat. Comput., vol. 5, no. 3, pp. 735-743, Jan. 1984

\bibitem{SVM}
L. Jiao, L. Bo, and L. Wang, "Fast sparse approximation for least squares
support vector machine," IEEE Trans. Neural Netw., vol. 18, no. 3,
pp. 685-697, May 2007.

\bibitem{KRR}
S. An, W. Liu, and S. Venkatesh, "Face recognition using kernel ridge
regression," in Proc. IEEE Comput. Soc. Conf. Comput. Vis. Pattern
Recognit., Minneapolis, MN, USA, pp. 1-8, Jun. 2007.

\bibitem{WLSR}
T. Strutz, "Data Fitting and Uncertainty: A Practical Introduction to
Weighted Least Squares and Beyond," Wiesbaden, Germany: Vieweg,
2010.


\bibitem{SRC}
J. Wright, A.Y. Yang, A. Ganesh, et al, "Robust face recognition via
sparse representation," IEEE Trans. Pattern Anal. Mach. Intell., vol. 31,
no. 2, pp. 210-227, 2009.

\bibitem{LRC}
I. Naseem, R. Togneri, and M. Bennamoun, "Linear regression for face
recognition," IEEE Trans. Pattern Anal. Mach. Intell., vol. 32, no. 11, pp.
2106-2112, 2010.


\bibitem{CRC}
L. Zhang, M. Yang, and X. Feng, "Sparse representation or collaborative
representation: Which helps face recognition?" in Proc. of IEEE Int. Conf.
Comput. Vis., pp. 471-478, 2011.

\bibitem{ProCRC}
S. Cai, L. Zhang, W. Zuo, et al. "A probabilistic collaborative representation based approach for pattern classification," in Proc. of IEEE Conf.
Comput. Vis. Pattern Recognit., pp. 2950-2959, 2016.

\bibitem{DLSR}
S. M. Xiang, F. P. Nie, G. F. Meng, C. H. Pan, and C. S. Zhang,
"Discriminative least squares regressions for multiclass classification and
feature selection," IEEE Trans. Neural Netw. Learn. Syst., vol. 23, no. 11,
pp. 1738-1754, Nov. 2012.

\bibitem{ReLSR}
X.-Y. Zhang, L. Wang, S. Xiang, and C.-L. Liu, "Retargeted least squares
regression algorithm," IEEE Trans. Neural Netw. Learn. Syst., vol. 26,
no. 9, pp. 2206-2213, Sep. 2015.

\bibitem{GReLSR}
L. Wang and C. Pan, "Groupwise retargeted least-squares regression,"
IEEE Trans. Neural Netw. Learn. Syst., vol. 29, no. 4, pp. 1352-1358,
Apr. 2018.


\bibitem{RLSL}
X. Z. Fang, S. H. Teng, Z. H. Lai, et al. "Robust latent subspace
learning for image classification," IEEE transactions on neural networks
and learning systems, 29(6): 2502-2515, 2018.


\bibitem{ADMM}
S. Boyd, N. Parikh, E. Chu, B. Peleato, and J. Eckstein, "Distributed optimization
and statistical learning via the alternating direction method of multipliers," Found. Trends
Mach. Learn., vol. 3, no. 1, pp. 1122, 2011.

\bibitem{SVD}
J. F. Cai, E. J. Candes, and Z. Shen, "A singular value thresholding algorithm for matrix completion," SIAM J. Optimization, vol. 20,
no. 4, pp. 1956–1982, 2010.


\bibitem{IMSC}
J. Wen, Y. Xu, and H. Liu, "Incomplete Multiview Spectral Clustering with Adaptive Graph Learning," IEEE Transactions on Cybernetics, 2018. doi: 10.1109/TCYB.2018.2884715.


\bibitem{AR}
 A. M. Martinez and R. Benavente, "The AR face database," CVC,
New Delhi, India, Tech. Rep. 24, 1998.

\bibitem{CMU_PIE}
T. Sim, S. Baker, M. Bsat, "The CMU pose, illumination, and expression
(PIE) database," in Proc. of IEEE Int. Conf. Autom. Face Gesture
Recognit., pp. 46-51, 2002.

\bibitem{Feret}
P. J. Phillips, H. Moon, S. A. Rizvi, and P. J. Rauss, "The FERET
evaluation methodology for face-recognition algorithms," IEEE Trans.
Pattern Anal. Mach. Intell., vol. 22, no. 10, pp. 1090-1104, Oct. 2000.


\bibitem{COIL}
 S. A. Nene, S. K. Nayar and H. Murase, "Columbia Object Image
Library (COIL-100)," Technical Report, CUCS-006-96, 1996.



\bibitem{p}
X. Cai, C. Ding, F. Nie, and H. Huang, "On the equivalent of low-rank
linear regressions and linear discriminant analysis based regressions," in
Proc. 19th ACM SIGKDD Conf. Knowl. Discovery Data Mining, pp. 1124-1132, 2013.

\bibitem{t-SNE}
Y.-L. Boureau, F. Bach, Y. LeCun, and J. Ponce, "Learning mid-level
features for recognition," in Proc. 23rd IEEE Conf. Comput. Vis. Pattern
Recognit., San Francisco, CA, USA, pp. 2559-2566, Jun. 2010.



\end{thebibliography}
\end{document}